\documentclass[10pt,twocolumn,letterpaper]{article}

\usepackage{wacv}
\usepackage{times}
\usepackage{epsfig}
\usepackage{graphicx}
\usepackage{amsmath}
\usepackage{amssymb}
\usepackage{url}

\wacvfinalcopy 

\def\wacvPaperID{****} 

\ifwacvfinal\fi
\begin{document}

\title{Feature Perceptual Loss for Variational Autoencoder}

\author{Xianxu Hou \\
University of Nottingham, Ningbo, China\\
{\tt\small xianxu.hou@nottingham.edu.cn}
\and
Ke Sun \\
University of Nottingham, Ningbo, China\\
{\tt\small ke.sun@nottingham.edu.cn}
\and
Linlin Shen \\
Shenzhen University, Shenzhen, China\\
{\tt\small llshen@szu.edu.cn}
\and
Guoping Qiu \\
University of Nottingham, Ningbo, China\\
{\tt\small guoping.qiu@nottingham.edu.cn}
}

\maketitle
\ifwacvfinal\thispagestyle{empty}\fi

\begin{abstract}
   We consider unsupervised learning problem to generate images like Variational Autoencoder (VAE) and Generative Adversarial Network (GAN), which are two popular generative models around this problem. Recent works on style transfer have shown that higher quality images can be generated by optimizing feature perceptual loss, which is based on pretrained deep convolutional neural network (CNN). We propose to train VAE by using feature perceptual loss to measure the similarity between the input and generated images instead of pixel-by-pixel loss. Testing on face image dataset, our model can produce better qualitative results than other models. Moreover, our experiments demonstrate that the learned latent representation in our model has powerful capability to capture the conceptual and semantic information of natural images, and achieve state-of-the-art performance in facial attribute prediction.

\end{abstract}

\section{Introduction}
Deep Convolutional Neural Networks (CNNs) have been used to achieve state-of-the-art performances in many supervised computer vision tasks such as image classification \cite{krizhevsky2012imagenet,simonyan2014very}, retrieval \cite{babenko2014neural}, detection \cite{girshick2014rich,sermanet2013overfeat}, and captioning \cite{karpathy2015deep, vinyals2015show}. Deep CNNs-based generative models, a branch of unsupervised learning techniques in machine learning, have become a hot research topic in computer vision area in recent years. A generative model trained with a given dataset can be used to generate data like the samples in the dataset, learn the internal essence of the dataset and "store" all the information in the limited parameters that are significantly smaller than the training dataset. 


Variational Autoencoder (VAE) \cite{kingma2013auto,rezende2014stochastic} has become a popular generative model, allowing us to formalize this problem in the framework of probabilistic graphical models with latent variables. By default, pixel-by-pixel measurement like L2 loss, or logistic regression loss is used to measure the difference between reconstructed and original images. Such measurements are easily implemented and effective for deep neural network training. However, the generated images are not clear and tend to be very blurry when compared to natural images. This is because the pixel-by-pixel loss is not good enough to capture the visual perceptual difference between two images and it is not the way how humans look at the world. For example, the same image offsetted by a few pixels has little visual perceptual difference for humans, but it could have very high pixel-by-pixel loss.

In this paper, we try to improve the standard (plain) VAE by replacing the pixel-by-pixel loss with feature perceptual loss which is the difference between high level features of images extracted from hidden layer in pretrained deep convolutional neural networks such as AlexNet \cite{krizhevsky2012imagenet} and VGGNet \cite{simonyan2014very} trained on ImageNet \cite{ILSVRC15}. The high-level feature-based loss has been successfully applied to deep neural network visualization \cite{simonyan2013deep,yosinski2015understanding}, texture synthesis and style transfer \cite{gatys2015neural,gatys2015texture}, demonstrating superiority over pixel-by-pixel loss. We also explore the conceptual representation capability of the learned latent space, and use it for facial attribute prediction.

\begin{figure*}
\begin{tabular}{ccc}
\rule{0pt}{1ex}\hspace{2.24mm}\includegraphics[width=16cm]{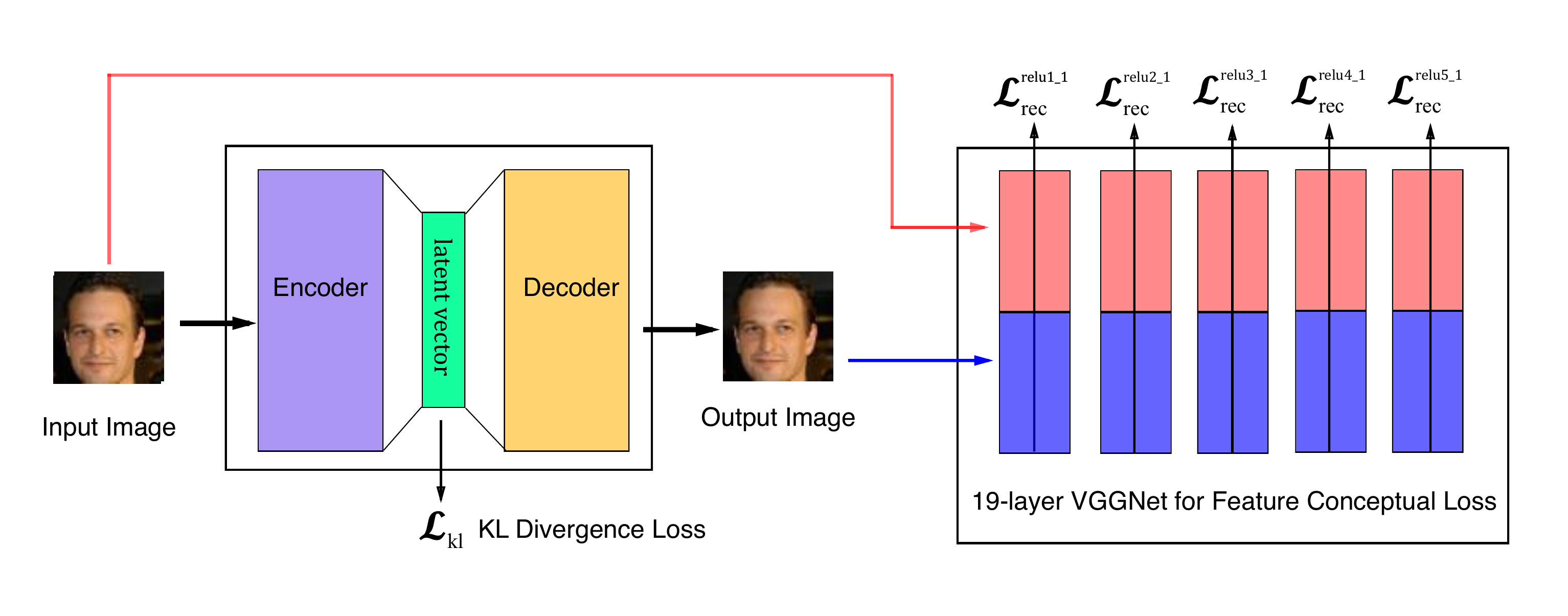}\\[-0.1pt]
\end{tabular}
\caption{Model Overview. The left is a deep CNN-based Variational Autoencoder, and the right is a pretrained deep CNN used to compute feature perceptual loss.}
\label{fig:overview}
\end{figure*}

\section{Related Work}
\textbf{Variational Autoencoder (VAE).}
A VAE \cite{kingma2013auto} helps us to do two things. Firstly it allows us to encode an image $x$ to a small dimension latent vector $z = Encoder(x) \sim q(z|x)$ with an encoder network, and then an decoder network is used to decode the latent vector $z$ back to an image that will be as similar as the original image $\bar{x} = Decoder(z) \sim p(x|z)$. That is to say, we need to maximize marginal log-likelihood of each observation (pixel) in x, and the VAE reconstruction loss $\mathcal{L}_{rec}$ is negative expected log-likelihood of observations in x. Another important property of VAE is able to control the distribution of latent vector $z$, which has characteristic of being independent unit Gaussian random variables, i.e., $z \sim \mathcal{N}(0, I)$. Moreover, the difference between the distribution of $q(z|x)$ and the distribution of a Gaussian distribution (called KL Divergence) can be quantified and minimized using gradient descent algorithm \cite{kingma2013auto}. Therefore, VAE models can be trained by optimizing the sum of the reconstruction loss ($\mathcal{L}_{rec}$) and KL divergence loss ($\mathcal{L}_{kl}$) using gradient descent.
$$\mathcal{L}_{rec} = - \mathbb{E}_{q(z|x)} [\log p(x|z)]$$ 
$$\mathcal{L}_{kl} = D_{kl}(q(z|x)||p(z)) $$ 
$$ \mathcal{L}_{vae} = \mathcal{L}_{rec} + \mathcal{L}_{kl}$$

Several methods have been proposed to improve the performance of VAE. \cite{kingma2014semi} extends the variational auto-encoders to semi-supervised learning with class labels, \cite{yan2015attribute2image} proposes a variety of attribute-conditioned deep variational auto-encoders, and demonstrates that they are capable of generating realistic faces with diverse appearance, Deep Recurrent Attentive Writer (DRAW) \cite{gregor2015draw} combines spatial attention mechanism with a sequential variational auto-encoding framework that allows iterative generation of images. Considering the shortcoming of pixel-by-pixel loss, \cite{ridgeway2015learning} replaces pixel-by-pixel loss with multi-scale structural-similarity score (MS-SSIM) and demonstrates that it can better measure human perceptual judgments of image quality. \cite{lamb2016discriminative} proposes to enhence the objective function with discriminative regularization. Another approach \cite{larsen2015autoencoding} tries to combine VAE and generative adversarial network (GAN) \cite{radford2015unsupervised,goodfellow2014generative}, and use the learned feature representation in the GAN discriminator as basis for the VAE reconstruction objective.

\begin{figure*}
\begin{tabular}{ccc}
\rule{0pt}{1ex}\hspace{2.24mm}\includegraphics[width=16cm]{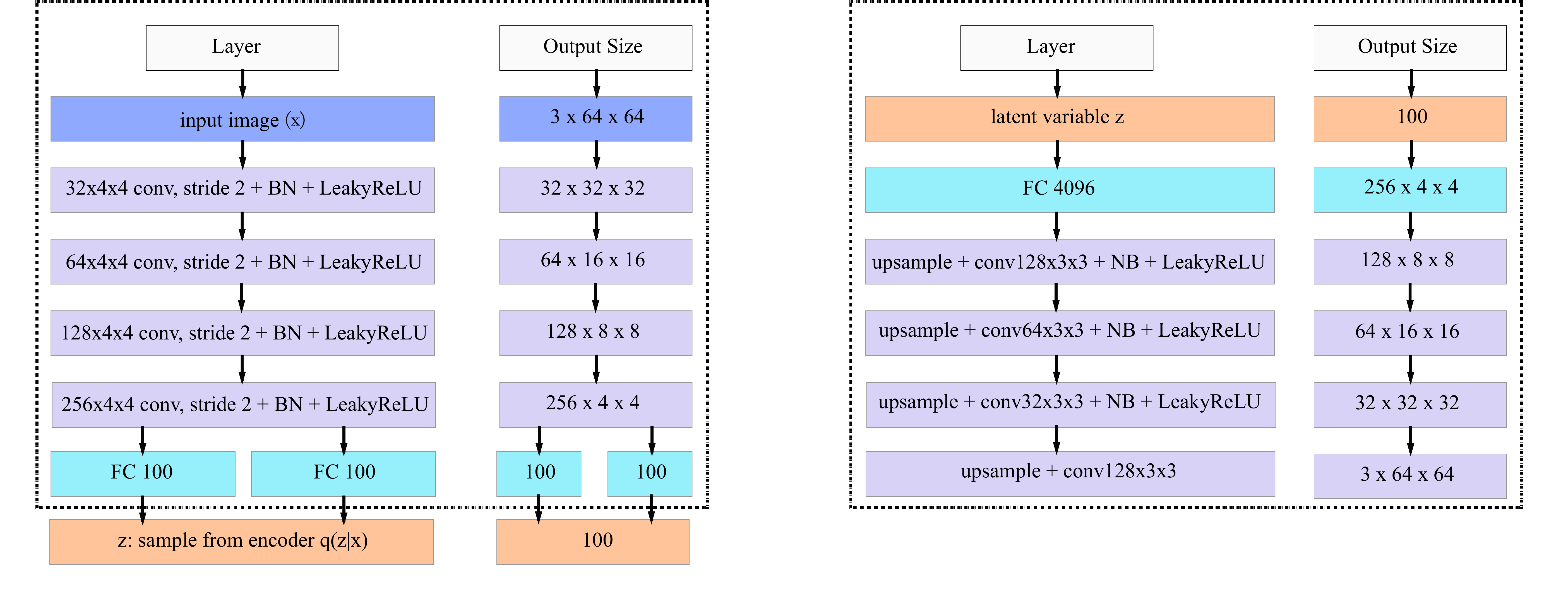}\\[-0.1pt]
\end{tabular}
\caption{Autoencoder network architecture. The left is encoder network, and the right is decoder network.}
\label{fig:autoencoder}
\end{figure*}

\textbf{high-level feature perceptual loss.}
Several recent papers successfully generate images by optimizing perceptual loss, which is based on the high-level features extracted from pretrained deep convolutional neural networks. Neural style transfer \cite{gatys2015neural} and texture synthesis \cite{gatys2015texture} tries to jointly minimize high-level feature reconstruction loss and style reconstruction loss by optimization. Additionally images can be also generated by maximizing classification scores or individual features \cite{simonyan2013deep,yosinski2015understanding}. Other works try to train a feed-forward network for real-time style transfer \cite{johnson2016perceptual,ulyanov2016texture,li2016combining} and super-resolution \cite{johnson2016perceptual} based on feature perceptual loss. In this paper, we train a deep convolutional variational autoencoder (CVAE) for image generation by replacing pixel-by-pixel reconstruction loss with high-level feature perceptual loss based on pre-trained network.

\section{Method}
Our system consists two main components as shown in Figure \ref{fig:overview}: an autoencoder network including an encoder network($E(x)$) and a decoder network($D(z)$), and a loss network ($\Phi$) that is a pretrained deep convolutional neural network to define feature perceptual loss. An input image $x$ is encoded as a latent vector $z = E(x)$, which will be decoded ($\bar{x} = D(z)$) back to image space. After training, new image can be generated by decoder network with a given vector $z$. In order to train a VAE, we need two loss functions, one is KL divergence loss ($\mathcal{L}_{kl} = D_{kl}(q(z|x)||p(z))$) \cite{kingma2013auto} which is used to make sure that the latent vector $z$ is an independent unit Gaussian random variable. The other is feature reconstruction loss. Instead of direct comparing the input image and the generated image in the pixel space, we pass both of them to a pre-trained deep convolutional neural network $\Phi$ respectively and then measure the difference between hidden layer representation, i.e., $\mathcal{L}_{rec} = \mathcal{L}^1 + \mathcal{L}^2 + ... + \mathcal{L}^l$, where $\mathcal{L}^l$ represents the feature loss at the $l^{th}$ hidden layer. Thus, we use the high-level feature loss to better measure perceptual and semantic differences between the two images, this is because the pretrained network on image classification has already incorporated perceptual and semantic information we desire for. During the training, the pretrained loss network is fixed and just for high-level feature extraction, and KL divergence loss $\mathcal{L}_{kl}$ is just used to update encoder network while the reconstruction feature loss $\mathcal{L}_{rec}$ is responsible for updating parameters of both encoder and decoder.

\subsection{Variational Autoencoder Network Architecture}
Both encoder and decoder network are based on deep convolutional neural network (CNN) like AlexNet \cite{krizhevsky2012imagenet} and VGGNet \cite{simonyan2014very}. We construct 4 convolutional layers in encoder network with 4 x 4 kernels, and the stride is fixed to be 2 to achieve spatial downsampling instead of using deterministic spatial functions such as maxpooling. Each convolutional layer is followed by a batch normalization layer and a LeakyReLU activation layer. Then two fully-connected output layers (for mean and variance) are added to encoder, and will be used to compute the KL divergence loss and sample latent variable $z$ (see \cite{kingma2013auto,Joost2015} for details). For decoder, we use 4 convolutional layers with 3 x 3 kernels and set stride to be 1, and replace standard zero-padding with replication padding, i.e., feature map of an input is padded with the replication of the input boundary. For upsampling we use nearest neighbor method by scale of 2 instead of fractional-strided convolutions used by other works \cite{long2015fully,radford2015unsupervised}. We also use batch normalization to help stabilize training and use LeakyReLU as activation function. The details of autoencoder network architecture is shown in Figure \ref{fig:autoencoder}.

\subsection{Feature Perceptual Loss}

Feature perceptual loss of two images is defined as the difference between the hidden features in a pretrained deep convolutional neural network $\Phi$. Similar to \cite{gatys2015neural}, we use VGGNet \cite{simonyan2014very} as the loss network in our experiment, which is trained for classification problem on ImageNet dataset. The core idea of feature perceptual loss is to seek the similarity between the hidden representation of two images, and the input images tend to be similar from perceptual and semantic aspect if the difference of hidden representation is small. Specifically, let $\Phi(x)^l$ denote the representation of a $l^{th}$ hidden layer when input image $x$ is fed to network $\Phi$. Mathematically $\Phi(x)^l$ is a 3D volume block array of shape [$C^l$ x $W^l$ x $H^l$], where $C^l$ is the number of filters, $W^l$ and $H^l$ represent the width and height of each feature map for the $l^{th}$ layer. The feature perceptual loss for one layer ($\mathcal{L}^l_{rec}$) between two images $x$ and $\bar{x}$ can be simply defined by squared euclidean distance. Actually it is quite like pixel-by-pixel loss for images except that the color channel is not 3 any more. 
$$
\mathcal{L}^l_{rec} =  \frac{1}{2C^lW^lH^l}  \sum_{c=1}^{C^l}  \sum_{w=1}^{W^l}  \sum_{h=1}^{H^l} (\Phi(x)^l_{c,w,h} - \Phi(\bar{x})^l_{c,w,h})^2
$$

By optimization to reconstruct images from noise, \cite{gatys2015neural,johnson2016perceptual} show that reconstruction from lower layers is almost perfect. While using higher layers, pixel information such as color and shape are changed although overall spatial structures can be preserved. In our paper, our reconstruction loss is defined as the total loss at different layers of VGG Network, i.e., $\mathcal{L}_{rec} = \sum_l\mathcal{L}^l_{rec}$. Additionally we adopt the KL divergence loss $\mathcal{L}_{kl}$ \cite{kingma2013auto} to regularize the encoder network to control the distribution of latent variable $z$. To train VAE, we jointly minimize the KL divergence loss $\mathcal{L}_{kl}$ and feature perceptual loss $\mathcal{L}^l_{rec}$ for different layers, i.e.,
$$
\mathcal{L}_{total} = \alpha \mathcal{L}_{kl} + \beta \sum_i^l (\mathcal{L}^l_{rec})
$$

where $\alpha$ and $\beta$ are weighted parameters for KL Divergence and image reconstruction. It is quite similar to style transfer \cite{gatys2015neural} if we treat KL Divergence as style reconstruction.

\section{Experiments}
In this paper, we perform experiments on face images to test our method. Specifically we compare the performance of our model trained by high-level feature perceptual loss with other generative models. Furthermore, we also investigate the latent space to seek semantic relationship between different latent representation and apply it to facial attribute prediction.

\subsection{Training Details} 
Our model is trained on CelebFaces Attributes (CelebA) Dataset \cite{liu2015deep}. CelebA is a large-scale face attributes dataset with 202,599 number of face images, and 5 landmark locations, 40 binary attributes annotations per image. We build the training dataset by cropping and scaling the alignment images to 64 x 64 pixels like \cite{larsen2015autoencoding,radford2015unsupervised}. We train our model with a batch size of 64 for 5 epochs over the training dataset and use Adam method for optimization \cite{kingma2014adam} with initial learning rate of 0.0005, which is decreased by 0.5 for the following epochs. The 19-layer VGGNet \cite{simonyan2014very} is chosen as loss network $\Phi$ to construct feature perceptual loss for image reconstruction. We experiment with different layer combinations to construct feature perceptual loss and report the results by using layers relu1\_2, relu2\_1, relu3\_1. In addition, the dimension of latent vector $z$ is set to be 100, and the loss weighted parameters $\alpha$ and $\beta$ are 1 and 0.8 respectively. Our implementation is built on deep learning framework Torch \cite{collobert2011torch7} and style transfer implementation \cite{Johnson2015}.

\subsection{Qualitative Results for Image Generation}
In this paper, we also train additional two generative models for comparison. One is the plain Variational Autoencoder (PVAE), which has the same architecture as our proposed model, but trained with pixel-by-pixel loss in the image space. The other is Deep Convolutional Generative Adversarial Networks (DCGAN) consisting of a generator and a discriminator network \cite{radford2015unsupervised}, which has shown the ability to generate high quality images from a noise vector. DCGAN is trained with open source code \cite{radford2015unsupervised} in Torch. The comparison is divided into two parts: arbitrary face images generated by decoder based on latent vector $z$ drawn from $\mathcal{N}(0, 1)$, and face image reconstruction.

\begin{figure}
\begin{tabular}{ccc}
\rule{0pt}{1ex}\hspace{2.24mm}\includegraphics[width=8cm]{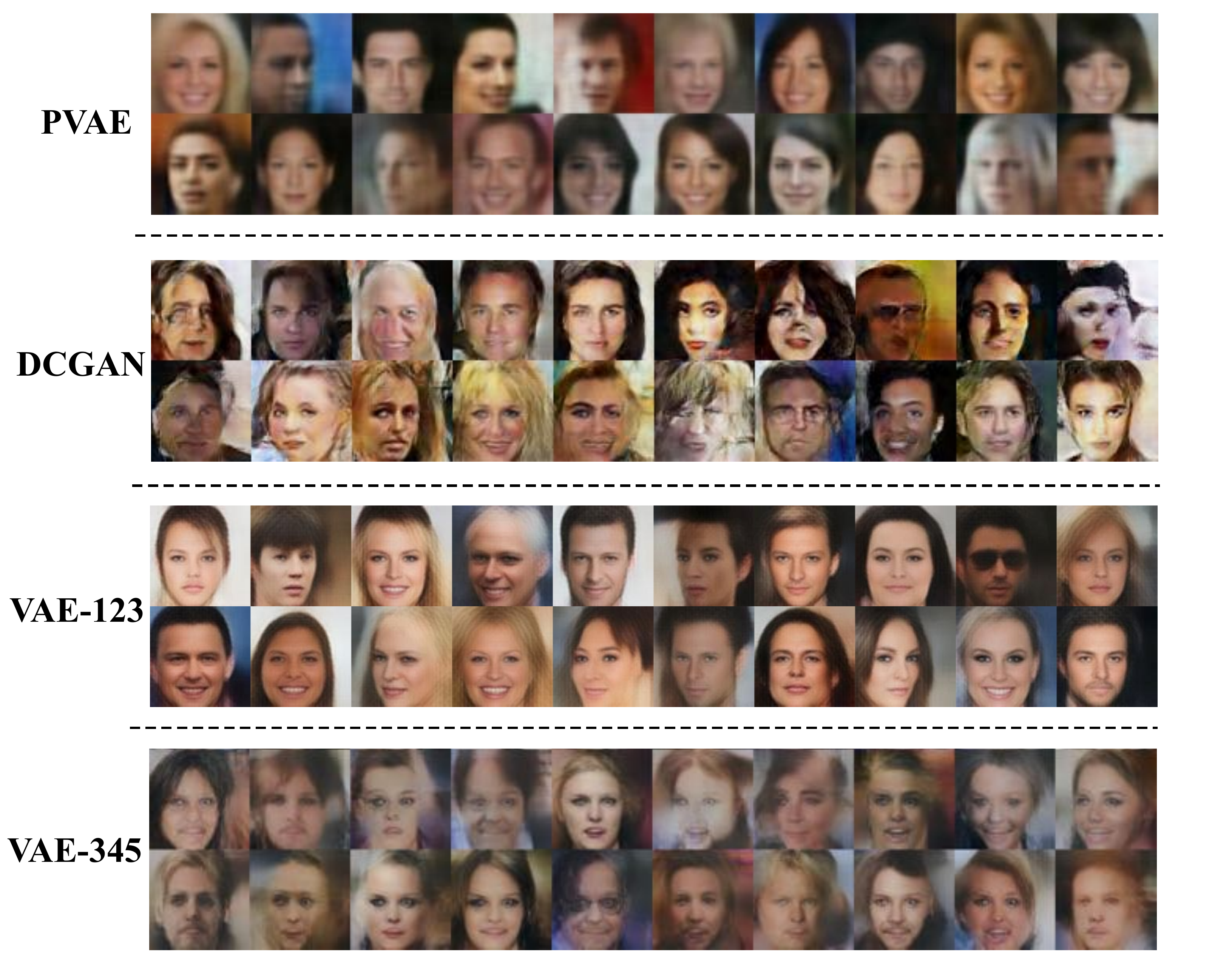}\\[-0.1pt]
\end{tabular}
\caption{Generated fake face images from 100-dimension latent vector $z \sim \mathcal{N}(0, 1)$ from different models. The first part is generated from decoder network of plain variational autoencoder (PVAE) trained with pixel-based loss \cite{kingma2013auto}, the second part is generated from generator network of DCGAN \cite{radford2015unsupervised}, and the third part is our method trained with feature perceptual loss.}
\label{fig:random_faces}
\end{figure}

In the first part, random face images (shown in Figure \ref{fig:random_faces}) are generated by three models from latent vector $z$ drawn from $\mathcal{N}(0, 1)$. We can see that the generated face images by plain VAE tend to very blurry, even though overall spatial face structure can be preserved. It is very hard for plain VAE to generate clear facial parts such as eyes and noses, this is because it tries to minimize the reconstruction difference between two images with pixel-by-pixel loss. The pixel-based loss is problematic due to no semantic and perceptual information contained. DCGAN can generate clean and sharp face images containing clearer facial textures, however it has the facial distortion problem and sometimes generates weird faces. Our method based on feature perceptual loss can achieve better results, generating faces of different genders, ages and races with clear noses and eyes. What's more, face images with sunglasses and white clean teeth can be also randomly generated. One problem found in our method is that the generated hair tends to be blurry in most samples, and we think it is because of the subtle texture of human hair.

\begin{figure}
\begin{tabular}{ccc}
\rule{0pt}{1ex}\hspace{2.24mm}\includegraphics[width=8cm]{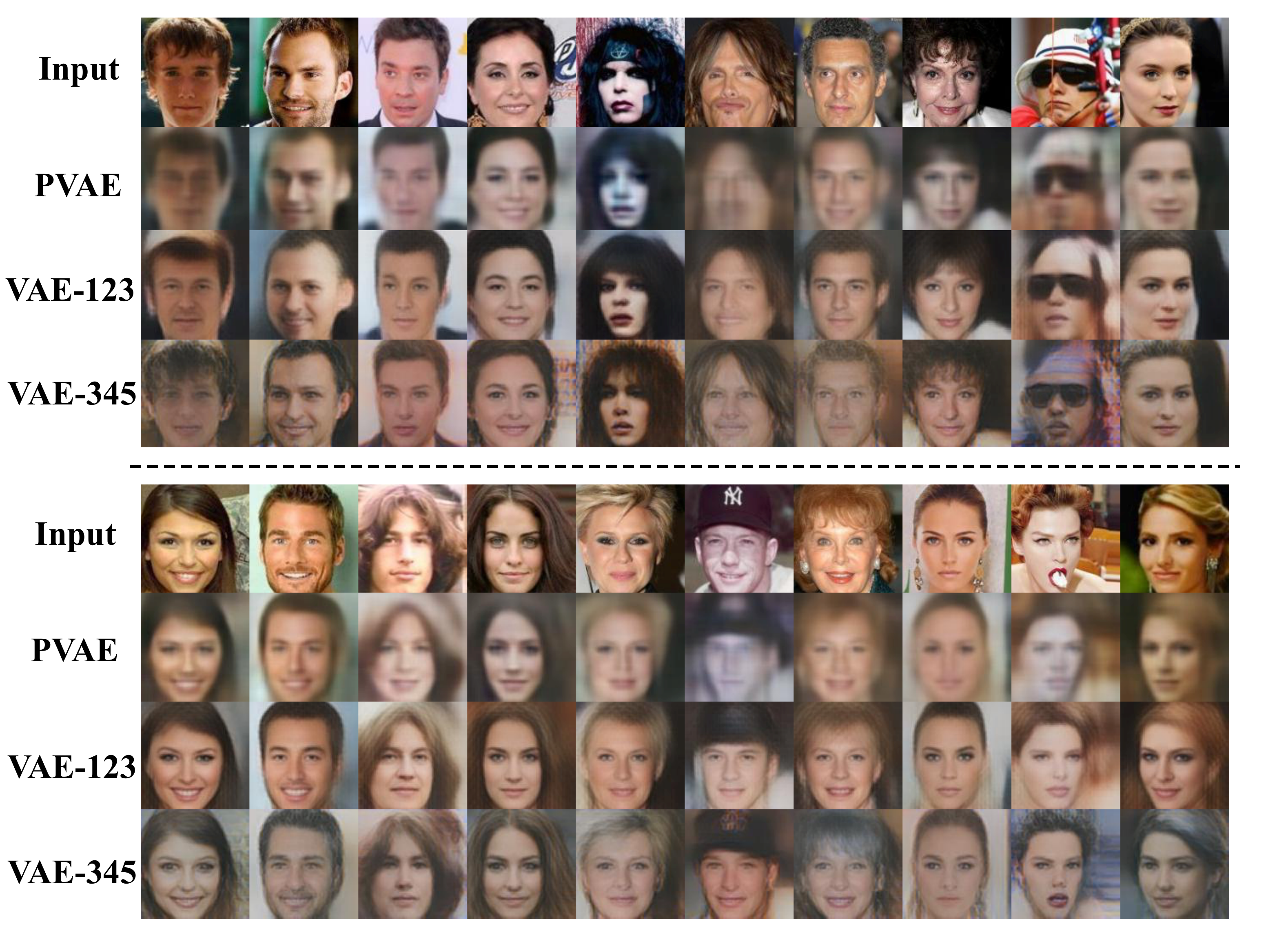}\\[-0.1pt]
\end{tabular}
\caption{Image reconstruction from different models. The first row is input image, the second row is generated from decoder network of plain variational autoencoder (PVAE) trained with pixel-based loss \cite{kingma2013auto}, and the last row is our method trained with feature perceptual loss.}
\label{fig:reconstruction}
\end{figure}

We also compare the reconstruction results (shown in Figure \ref{fig:reconstruction}) between plain VAE and our method, and DCGAN is not compared because of no input image in their model. We can get similar conclusion as above between two methods. Even though the reconstruction is not perfect and the generated face images tend to be blurry when compared to input images, our method is much better than plain VAE.

\subsection{Investigating Learned Latent Space}
\subsubsection{Linear interpolation of latent space}
In order to get a better understanding of what our model has learned, we investigate the property of the $z$ representation in the latent space from our encoder network, and the relationship between the different learned latent vectors.

\begin{figure}
\rule{0pt}{1ex}\hspace{2.24mm}\includegraphics[width=8cm]{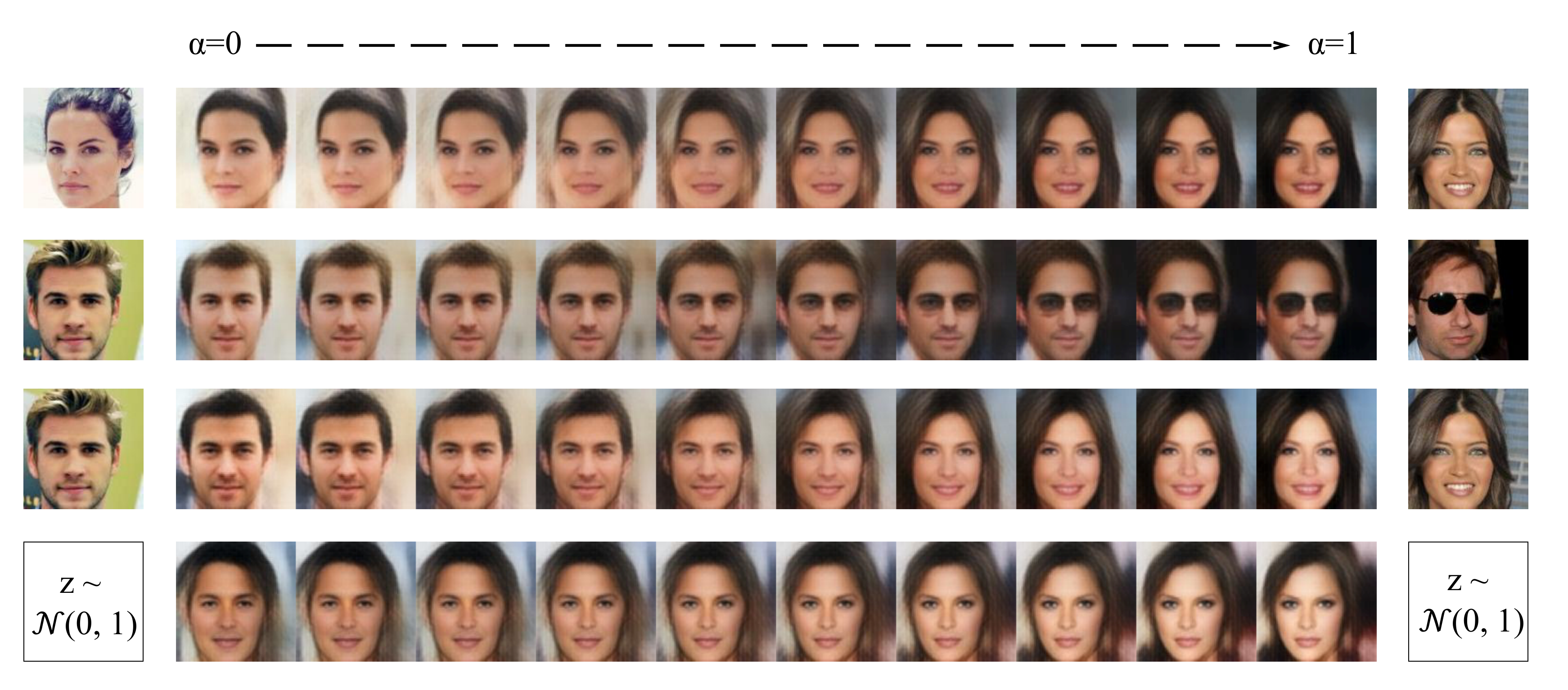}\\[-0.1pt]
\caption{Linear interpolation for latent vector. Each row is the interpolation from left latent vector $z_{left}$ to right latent vector $z_{right}$. e.g. $(1-\alpha) z_{left} + \alpha z_{right}$. The first row is transitions from a non-smiling woman to a smiling woman, the second row is transitions from a man without sunglass to a man with sunglass, the third row is transitions from a man to a woman, and the last row is transitions between two fake faces decoded from $z \sim \mathcal{N}(0, 1)$.}
\label{fig:linear_interpolation}
\end{figure}

As shown in Figure \ref{fig:linear_interpolation}, we investigate the generated images from two latent vectors denoted as $z_{left}$ and $z_{right}$. The interpolation is defined by linear transformation $z = (1-\alpha) z_{left} + \alpha z_{right}$, where $\alpha = 0, 0.1, \dots, 1$, and then $z$ is fed to decoder network to generate new face images. In this paper, we provide three examples for latent vector $z$ encoded from input images and one example for $z$ randomly drawn from $\mathcal{N}(0, 1)$. From the first row in Figure \ref{fig:linear_interpolation}, we can see the smooth transitions between $vector$("Woman without smiling and short hair") and $vector$("Woman with smiling and long hair"). Little by little the hair become longer, the distance between lips become larger and teeth is shown in the end for smiling, and pose turns from looking slightly left to looking front. Additionally we provide examples of transitions between $vector$("Man without sunglass") and $vector$("Man with sunglass"), and $vector$("Man") and $vector$("Woman").
\begin{figure}
\centering
\begin{tabular}{ccc}
\rule{0pt}{1ex}\hspace{2.24mm}\includegraphics[width=8cm]{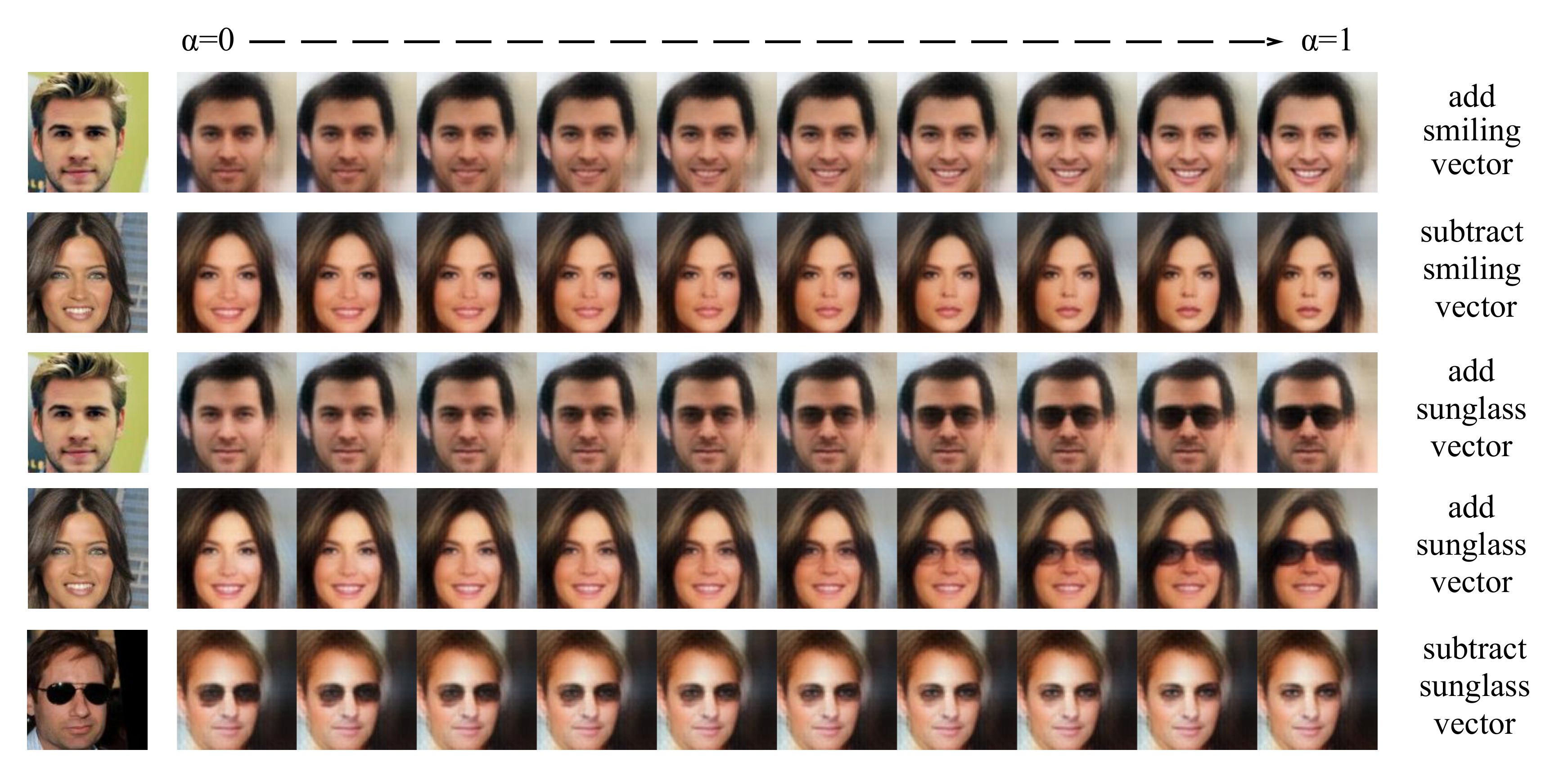}\\[-0.1pt]
\end{tabular}
\caption{Vector arithmetic for visual attributes. Each row is the generated faces from latent vector $z_{left}$ by adding or subtracting an attribute-specific vector. e.g. $z_{left}$ + $\alpha$ $z_{smiling}$, where $\alpha = 0, 0.1, \dots, 1$. The first row is the transitions by adding a smiling vector with a linear factor $\alpha$ from left to right, the second row is the transitions by subtracting a smiling vector, the third and fourth row are the results by adding a sunglass vector to latent representation for a man and women, and the last row shows results by the subtracting a sunglass vector.}
\label{fig:attribute_specific}
\end{figure}

\begin{figure}
\rule{0pt}{1ex}\hspace{2.24mm}\includegraphics[width=9cm]{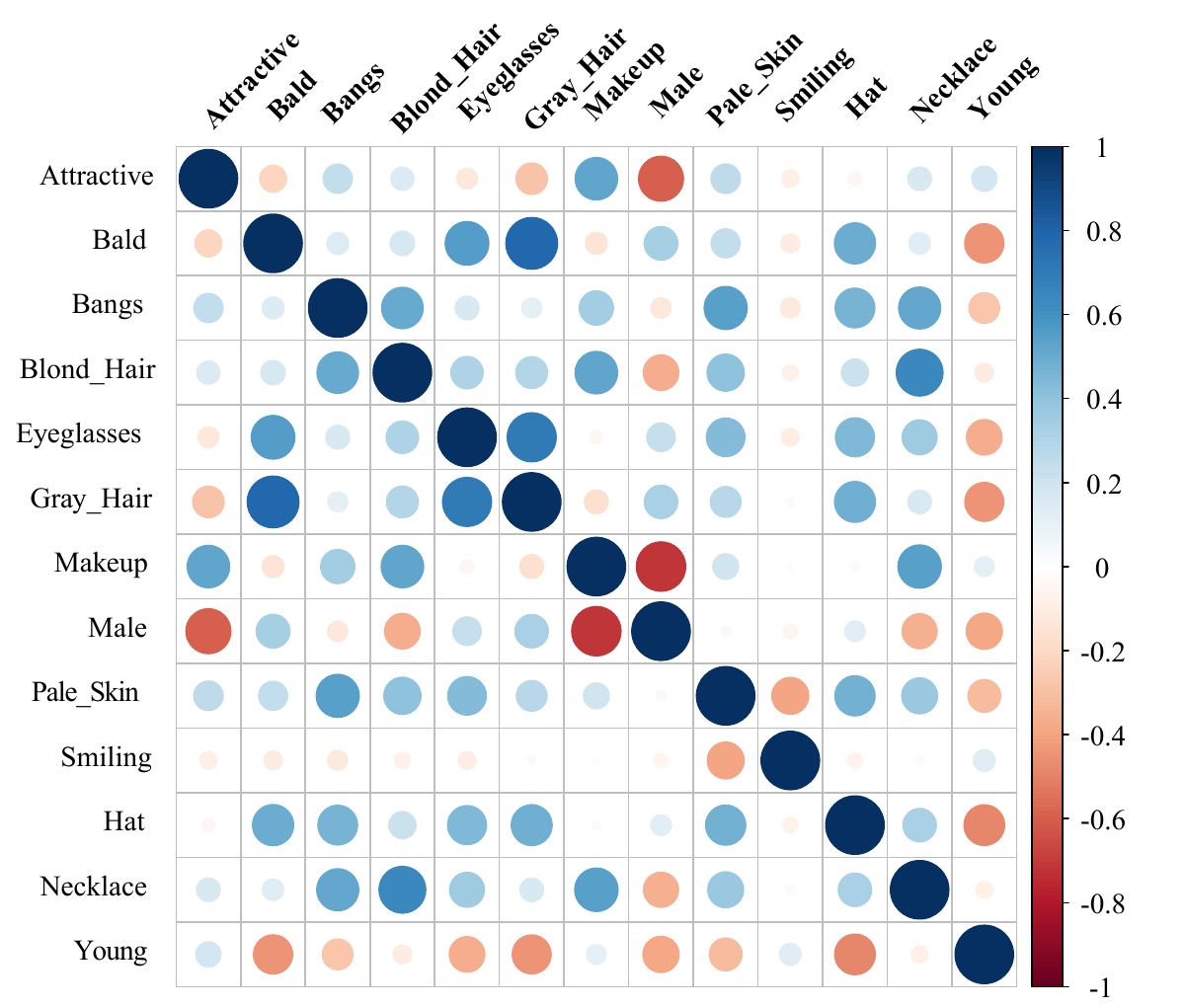}\\[-0.1pt]
\caption{Diagram for the correlation between selected facial attribute-specific vectors. The blue indicates positive correlation, while red represents negative correlation, and the color shades and sizes of the circle represent the strength the correlation.}
\label{fig:attr_embed_corr}
\end{figure}

\begin{figure*}
\begin{tabular}{ccc}
\rule{0pt}{1ex}\hspace{2.24mm}\includegraphics[width=18cm]{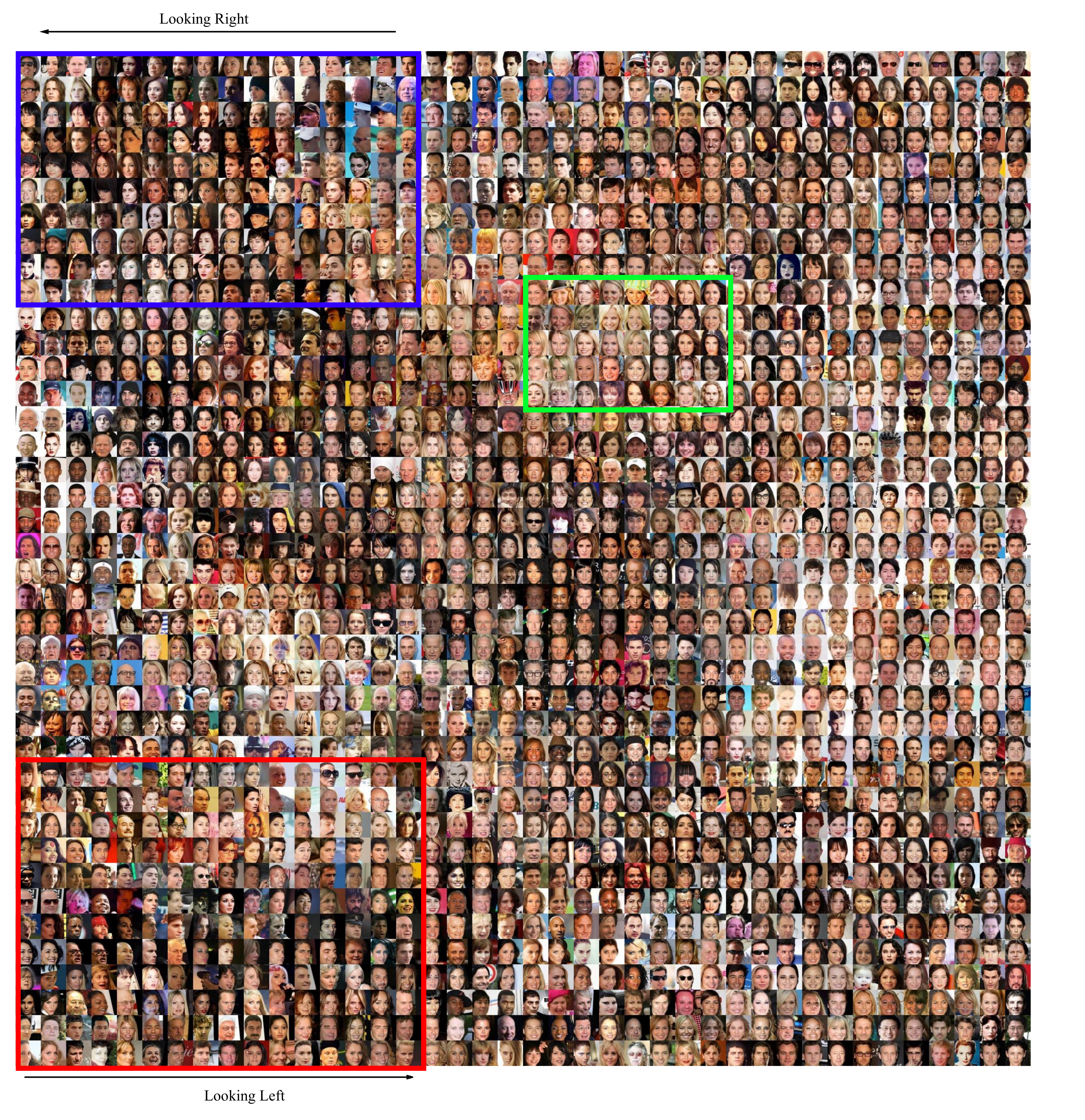}\\[-0.1pt]
\end{tabular}
\caption{Visualization of 400 x 400 face images by latent vectors with t-SNE algorithm \cite{maaten2008visualizing} }
\label{fig:cnn_embed_full_2k}
\end{figure*}

\begin{table*}
\footnotesize
\begin{center}
\begin{tabular}{ | l | l | l | l | l | l | l | l | l | l | l | l | l | l | l | l | l | l | l | l | l | l | }
\hline
    Method & \rotatebox[origin=c]{90}{5 Shadow} & \rotatebox[origin=c]{90}{Arch. Eyebrows} & \rotatebox[origin=c]{90} {Attractive} & \rotatebox[origin=c]{90} {Bags Un. Eyes} & \rotatebox[origin=c]{90} {Bald} & \rotatebox[origin=c]{90} {Bangs} & \rotatebox[origin=c]{90} {Big Lips} & \rotatebox[origin=c]{90} {Big Nose} & \rotatebox[origin=c]{90} {Black Hair} & \rotatebox[origin=c]{90} {Blond Hair} & \rotatebox[origin=c]{90} {Blurry} & \rotatebox[origin=c]{90} {Brown Hair} & \rotatebox[origin=c]{90} {Bushy Eyebrows} & \rotatebox[origin=c]{90} {Chubby} & \rotatebox[origin=c]{90} {Double Chin} & \rotatebox[origin=c]{90} {Eyeglasses} & \rotatebox[origin=c]{90} {Goatee} & \rotatebox[origin=c]{90} {Gray Hair} & \rotatebox[origin=c]{90} {Heavy Makeup} & \rotatebox[origin=c]{90} {H. Cheekbones} & \rotatebox[origin=c]{90} {Male} \\ \hline
    FaceTracer & 85 & 76 & 78 & 76 & 89 & 88 & 64 & 74 & 70 & 80 & 81 & 60 & 80 & 86 & 88 & 98 & 93 & 90 & 85 & 84 & 91 \\ \hline
    PANDA-w & 82 & 73 & 77 & 71 & 92 & 89 & 61 & 70 & 74 & 81 & 77 & 69 & 76 & 82 & 85 & 94 & 86 & 88 & 84 & 80 & 93 \\ \hline
    PANDA-l & 88 & 78 & \textbf{81} & 79 & 96 & 92 & 67 & 75 & 85 & 93 & 86 & 77 & 86 & 86 & 88 & 98 & 93 & 94 & 90 & 86 & 97 \\ \hline
    LNets+ANet & \textbf{91} & \textbf{79} & \textbf{81} & 79 & \textbf{98} & \textbf{95} & 68 & 78 & \textbf{88} & \textbf{95} & 84 & \textbf{80} & \textbf{90} & 91 & 92 & \textbf{99} & \textbf{95} & \textbf{97} & \textbf{90} & \textbf{87} & \textbf{98} \\ \hline
    VAE-Z & 89 & 77 & 75 & \textbf{81} & \textbf{98} & 91 & \textbf{76} & \textbf{79} & 83 & 92 & \textbf{95} & \textbf{80} & 87 & \textbf{94} & \textbf{95} & 96 & 94 & 96 & 85 & 81 & 90 \\ \hline
    VGG-FC & 83 & 71 & 68 & 73 & 97 & 81 & 51 & 77 & 78 & 88 & 94 & 67 & 81 & 93 & 93 & 95 & 93 & 94 & 79 & 64 & 84 \\ \hline
    \  & \  & \  & \  & \  & \  & \  & \  & \  & \  & \  & \  & \  & \  & \  & \  & \  & \  & \  & \  & \  & \  \\ \hline
    Method & \rotatebox[origin=c]{90}{Mouth S. O.} & \rotatebox[origin=c]{90}{Mustache} & \rotatebox[origin=c]{90}{Narrow Eyes} & \rotatebox[origin=c]{90}{No Beard} & \rotatebox[origin=c]{90}{Oval Face} & \rotatebox[origin=c]{90}{Pale Skin} & \rotatebox[origin=c]{90}{Pointy Nose} & \rotatebox[origin=c]{90}{Reced. Hairline} & \rotatebox[origin=c]{90}{Rosy Cheeks} & \rotatebox[origin=c]{90}{Sideburns} & \rotatebox[origin=c]{90}{Smiling} & \rotatebox[origin=c]{90}{Straight Hair} & \rotatebox[origin=c]{90}{Wavy Hair} & \rotatebox[origin=c]{90}{Wear. Earrings} & \rotatebox[origin=c]{90}{Wear. Hat} & \rotatebox[origin=c]{90}{Wear. Lipstick} & \rotatebox[origin=c]{90}{Wear. Necklace} & \rotatebox[origin=c]{90}{Wear. Necktie} & \rotatebox[origin=c]{90}{Young} & \rotatebox[origin=c]{90}{Average} & \  \\ \hline
    FaceTracer & 87 & 91 & 82 & 90 & 64 & 83 & 68 & 76 & 84 & 94 & 89 & 63 & 73 & 73 & 89 & 89 & 68 & 86 & 80 & 81.13 & \  \\ \hline
    PANDA-w & 82 & 83 & 79 & 87 & 62 & 84 & 65 & 82 & 81 & 90 & 89 & 67 & 76 & 72 & 91 & 88 & 67 & 88 & 77 & 79.85 & \  \\ \hline
    PANDA-l & \textbf{93} & 93 & 84 & 93 & 65 & 91 & 71 & 85 & 87 & 93 & \textbf{92} & 69 & 77 & 78 & 96 & \textbf{93} & 67 & 91 & 84 & 85.43 & \  \\ \hline
    LNets+ANet & 92 & 95 & 81 & \textbf{95} & 66 & 91 & 72 & 89 & 90 & \textbf{96} & \textbf{92} & 73 & \textbf{80} & \textbf{82} & \textbf{99} & \textbf{93} & 71 & \textbf{93} & \textbf{87} & \textbf{87.30} & \  \\ \hline
    VAE-Z & 80 & \textbf{96} & \textbf{89} & 88 & \textbf{73} & \textbf{96} & \textbf{73} & \textbf{92} & \textbf{94} & 95 & 87 & \textbf{79} & 74 & \textbf{82} & 96 & 88 & \textbf{88} & \textbf{93} & 81 & 86.95 & \  \\ \hline
    VGG-FC & 60 & 93 & 87 & 84 & 66 & 96 & 58 & 86 & 93 & 85 & 65 & 68 & 70 & 49 & 98 & 82 & 87 & 89 & 74 & 79.85 & \  \\ \hline
\end{tabular}
\end{center}
\caption{Performance comparison of 40 facial attributes prediction. The accuracies of FaceTracer \cite{kumar2008facetracer}, PANDA-w \cite{zhang2014panda}, PANDA-l \cite{zhang2014panda}, and LNets+ANet \cite{liu2015deep} are collected from \cite{liu2015deep}. PANDA-l, VAE-Z and VGG-FC use the truth landmarks to get the face part.}
\label{tab:performance_attribute_prediction}
\end{table*}


\subsubsection{Facial attributes manipulation}
The experiments above demonstrate interesting smooth transition's property between two learned latent vectors. In this part, instead of manipulating the overall face images, we seek to find a way to control a specific attribute of face images. In previous works, \cite{mikolov2013distributed} shows that $vector$("King") - $vector$("Man") + $vector$("Woman") generates a vector whose nearest neighbor was the $vector$("Queen") when evaluating learned representation of words. \cite{radford2015unsupervised} demonstrates that visual concepts such as face pose and gender could be manipulated by simple vector arithmetic. In this paper, we investigate two facial attributes wearing sunglass and smiling. We randomly choose 1000 face images with sunglass and 1000 without sunglass respectively from the CelebA dataset \cite{liu2015deep}, finally the two type of images are fed to our encoder network to compute the latent vectors, and the mean latent vectors are calculated for each type respectively, denoted as $z_{pos\_sunglass}$ and $z_{neg\_sunglass}$. We then define the difference $z_{pos\_sunglass} - z_{neg\_sunglass}$ as sunglass-specific latent vector $z_{sunglass}$. In the same way, we calculate the smiling-specific latent vector $z_{smiling}$. Then we apply the two attribute-specific vectors to different latent vectors $z$ by simple vector arithmetic, for instance, $z$ + $\alpha$ $z_{smiling}$. From Figure \ref{fig:attribute_specific}, by adding a smiling vector to the latent vector of a non-smiling man, we can observe the smooth transitions from non-smiling face to smiling face (the first row). What's more, the smiling appearance becomes more obvious when the factor $\alpha$ is bigger, while other facial attributes are able to remain unchanged. The other way round, when the latent vector of smiling woman is subtracted by the smiling vector, the smiling face can be translated to not smiling by only changing the shape of mouth (the second row in Figure \ref{fig:attribute_specific}). Moreover, we could add or wipe out a sunglass by playing with the calculated sunglass vector.

\subsubsection{Correlation between attribute-specific vectors}
Considering the conceptual relationship between different facial attributes in natural images, for instance, bald and gray hair are often related old people, we selected 13 of 40 attributes from CelebA dataset and calculate the attribute-specific vector respectively (the calculation is the same as calculating sunglass-specific vector above). We then visualize the correlation as shown in Figure \ref{fig:attr_embed_corr}, and the results are well consistent with human interpretation. We can see that $Attractive$ has a strong positive correlation with $Makeup$, and a negative correlation with $Male$ and $Gray\: Hair$. It makes sense that female is generally considered more attractive than male and uses a lot of makeup. Similarly, $Bald$ has a positive correlation with $Gray\:Hair$ and $Eyeglasses$, and a negative correlation with $Young$. Additionally, $Smiling$ seems to have no correlation with most of other attributes and only have a weak negative correlation with $Pale\:Skin$. It could be explained that $Smiling$ is a very common human facial expression and it could have a good match with many other attributes.

\subsubsection{Visualization of latent vectors}
Considering that the latent vectors are nothing but the encoding representation of the natural face images, we think that it may be interesting to visualize the natural images based on the similarity of the latent representation in an unsupervised way. Specifically we randomly choose 1600 face images from CelebA dataset and extract the corresponding 100-dimensional latent vectors, which are then reduced to 2-dimensional embedding by using t-SNE algorithm \cite{maaten2008visualizing}. t-SNE can arrange images that have a similar high-dimensional code (L2 distance) nearby in the embedding space. The visualization of 400 x 400 images is shown in Figure \ref{fig:cnn_embed_full_2k}, and we can discover that images with similar background (black or white) tend to be clustered as a group, and female with smiling can be clustered together (green rectangle in Figure \ref{fig:cnn_embed_full_2k}). What's more, the face pose information can also be captured even no pose annotations in the dataset. The face images in the upper left (blue rectangle) tend to look left and samples in the lower left (red rectangle) tend to look right, while in other area tend to look front.

\subsubsection{Facial attribute prediction}
In the end, we evaluate our model by applying latent vector to facial attribute prediction, which is a very challenging problem due to complex face variations. Similar to \cite{liu2015deep}, 20,000 images from CelebA dataset are selected for testing and the rest for training. Firstly we use ground truth landmark points to crop out the face parts of the original images like PANDA-l \cite{zhang2014panda}, and the cropped face images are fed to our encoder network to extract latent vectors, which are then used to train standard Linear SVM \cite{scikit-learn} classifiers. As a result, we train 40 binary classifiers for each attribute in CelebA dataset respectively. As a baseline, we also train different Linear SVM classifiers with 4096-dimensional deep features extracted from the last fully connected layer of pretrained VGGNet \cite{simonyan2014very}. We then compare our method with other state-of-the-art methods. The average of prediction accuracies of FaceTracer \cite{kumar2008facetracer}, PANDA-w \cite{zhang2014panda}, PANDA-l \cite{zhang2014panda}, and LNets+ANet \cite{liu2015deep} are 81.13, 79.85, 85.43 and 87.30 percent respectively. Our method with latent vector of VAE (VAE-Z) and VGG last layer features (VGG-FC) are 86.95 and 79.85 respectively. From Table \ref{tab:performance_attribute_prediction}, we can see that our method is comparable to the LNets+ANet and outperforms other methods. Our method can do a better job to predict $Wearing\_Necklace$, $Receding\_Hairline$ and $Pale\_Skin$. In addition, we notice that all the methods can achieve a good performance to predict $Bald$, $Wearing\_Hat$ and $Eyeglasses$, while they are very difficult to correctly predict attributes like $Big\_Lips$ and $Oval\_Face$. The reason we think is that attributes like whether wearing hat and eyeglasses or not are much more obvious in natural face images, than attributes whether having big lips and Oval face or not, and the extracted features are not able to capture such subtle differences. Future work is needed to find a way to extract better features which can also capture tiny variation of facial attributes.

\subsection{Discussion}
For (variational) autoencoder models, one essential part is to define a reconstruction loss to measure the similar between input image and generated image. The plain VAE adopts the pixel-by-pixel distance, which is problematic and the generated images tend to be very blurry. Inspired by the state-of-the-art works on style transfer and texture synthesis \cite{gatys2015neural,johnson2016perceptual,ulyanov2016texture}, we measure the reconstruction loss in VAE by feature perceptual loss based on pretrained deep convolutional neural networks (CNNs). Our experiments above have shown that feature perceptual loss can be used to improve the performance of VAE to generate high quality images. One explanation is that the hidden representation in a pretrained deep CNN could capture conceptual and semantic information of a given image since it has the ability to do classification, which is a human understanding task. Another benefit of using deep CNNs is that we can combine different level of hidden representation, which can provide more constraints for the reconstruction. Actually we could explore different combinations even add weights to different level representation to generate weird but interesting images. However, the feature perceptual loss is not perfect, the trained model fails to generate clear hair texture in our experiments even though it can do a good job for eyes, noses and mouths generation. For further work, trying to construct better reconstruction loss to measure the similarity of the output images and ground-truth images is essential for this problem. One possibility is to combine feature perceptual loss with generative adversarial networks(GAN).

The more interesting part of VAE is the linear structure in the learned latent space. Different images generated by decoder can be smoothly transformed to each other by simply linear combination of their latent vectors. Additionally attribute-specific latent vectors could be also calculated by encoding the annotated images and used to manipulate the related attribute of a given image while keeping other attributes unchanged, what's more, the correlation between attribute-specific vectors is well consistent with human understanding. Our experiments shows that the learned latent space of VAE can learn powerful representation of conceptual and semantic information of natural images, and it could be used for other applications like face attribute prediction.

\section{Conclusion}
In this paper, we try to improve the performance of image generation of VAE by combining feature perceptual loss based on pretrained deep CNNs to measure the similar of two images. We apply our model on face images and achieve comparable and better performance compared to different generative models (plain VAE and GAN). In addition, we fully explore the learned latent representation in our model and demonstrates it has powerful capability to capture the conceptual and semantic information of natural images. We also achieved state-of-the-art performance of facial attribute prediction based on the learned latent representation.

{\small
\bibliographystyle{ieee}
\bibliography{egbib}
}

\end{document}